\documentclass[conference]{IEEEtran}
\usepackage{graphicx}
\usepackage{amsmath}
\usepackage{cite}
\usepackage{url}
\usepackage{stfloats}

\begin{document}

\title{BiDexHand: Design and Evaluation of an Open-Source 16-DoF Biomimetic Dexterous Hand}
\author{
\IEEEauthorblockN{Zhengyang Kris Weng\IEEEauthorrefmark{1}}
\IEEEauthorblockA{\IEEEauthorrefmark{1}Center for Robotics and Biosystems, Northwestern University, Evanston, IL, USA\\
Email: zkweng@u.northwestern.edu}
}
\maketitle

\begin{abstract}
Achieving human-level dexterity in robotic hands remains a fundamental challenge for enabling versatile manipulation across diverse applications. This extended abstract presents BiDexHand, a cable-driven biomimetic robotic hand that combines human-like dexterity with accessible and efficient mechanical design. The robotic hand features 16 independently actuated degrees of freedom and 5 mechanically coupled joints through novel phalange designs that replicate natural finger motion. Performance validation demonstrated success across all 33 grasp types in the GRASP Taxonomy, 9 of 11 positions in the Kapandji thumb opposition test, a measured fingertip force of 2.14\,N, and the capability to lift a 10\,lb weight. As an open-source platform supporting multiple control modes including vision-based teleoperation, BiDexHand aims to democratize access to advanced manipulation capabilities for the broader robotics research community.
\end{abstract}


\section{Introduction}
The human hand is often regarded as the epitome of dexterous manipulation. Owing to its complex tendon-muscle architecture and high number of degrees of freedom (DoF), replicating its functionality in robotic systems presents significant challenges in mechanical design, actuation, and control. While prior work has made notable progress in developing multi-fingered dexterous hands \cite{bridgwater2012robonaut}\cite{kim2021ilda}\cite{shadowhand_patent}, some designs simplify mechanical complexity by limiting the number of actuated DoF \cite{bridgwater2012robonaut}, thereby compromising dexterity. Others maintain high DoF but are prohibitively expensive \cite{shadowrobot_hand_cost}, limiting their accessibility for broader research and application.

This work presents a cable-driven, biomimetic robotic hand that achieves rich, human-like motion through a low-cost design and accessible components. The human hand is typically modeled with 27 DoF, including a 6-DoF wrist and 21 DoF across the fingers \cite{hand_dof}. The proposed design achieves high finger-level dexterity with 16 independently actuated DoF and 5 mechanically coupled joints, resulting in 21 joints in total. When integrated with a 6-DoF robotic arm, the system achieves full spatial dexterity.

Key contributions of this work include: (1) a novel phalanx structure with a tendon-routing mechanism enabling biomimetic movement; (2) a ROS2-based control framework that supports multiple control input methods and teleoperation; and (3) a fully open-source, low-cost dexterity platform designed for integration with widely used robotic arms. By prioritizing affordability, modularity, and extensibility, this design aims to democratize access to multi-fingered dexterity and enable a broader range of researchers and developers to explore advanced manipulation capabilities.

\begin{figure}[!t]
    \centering
    \includegraphics[width=0.45\textwidth]{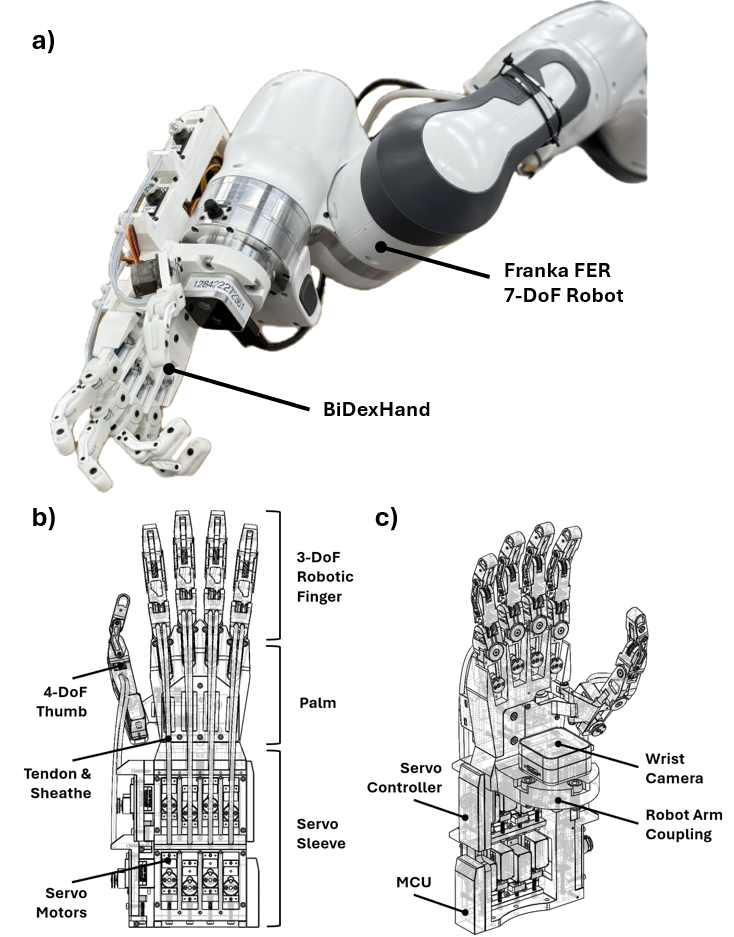}
    \caption{Overview of the BiDexHand. a) Integration of BiDexHand with the Franka FER Robot arm. b) Configuration of the BiDexHand includes four 3-DoF robotic fingers and a 4-DoF robotic thumb. c) Palmer view of the BiDexHand. The proposed design includes a coupling for Franka FER Robot Arm, and a RealSense D405 Wrist Camera.}
    \label{fig:overview}
\end{figure}

\section{Design and Methodology} 
\subsection{Phalanx Design}
\begin{figure}[h]
    \centering
    \includegraphics[width=0.45\textwidth]{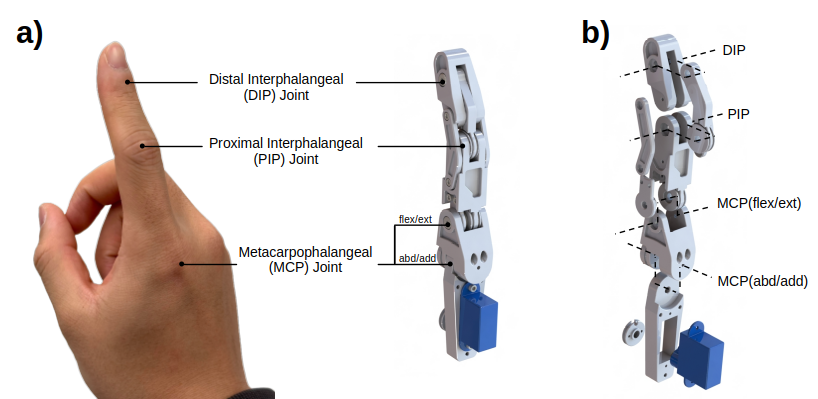}
    \caption{Configuration of the BiDexHand phalanx. a) Configuration of the robotic hand phalanx. b) Exploded view of the robotic phalanx.} 
    \label{fig:hand_comparison}
\end{figure}

The proposed robotic hand incorporates an anthropomorphic phalanx architecture, with each finger featuring four joints and offering three degrees of freedom. From the distal end, the joints include the distal interphalangeal (DIP), proximal interphalangeal (PIP), and metacarpophalangeal (MCP) joints, as illustrated in Figure~\ref{fig:hand_comparison}. Unlike in the human finger, the robotic MCP joint is functionally decoupled into two separate joints: one for abduction/adduction and another for flexion/extension.

In human fingers, the motion of the DIP and PIP joints is naturally coupled \cite{LEIJNSE20102381PIPKINEMATIC}, and the DIP versus PIP trajectories (DvPT) can be modeled analytically using a sigmoid function \cite{van2015analytical}. To replicate this biomechanical coupling while minimizing actuator count, the design employs a crossed four-bar linkage to passively drive the DIP joint in coordination with the actuated PIP joint. Specifically, the linkage is configured as an anti-parallelogram, enabling a closed-form relationship between joint angles.

\begin{figure}[h]
    \centering
    \includegraphics[width=0.45\textwidth]{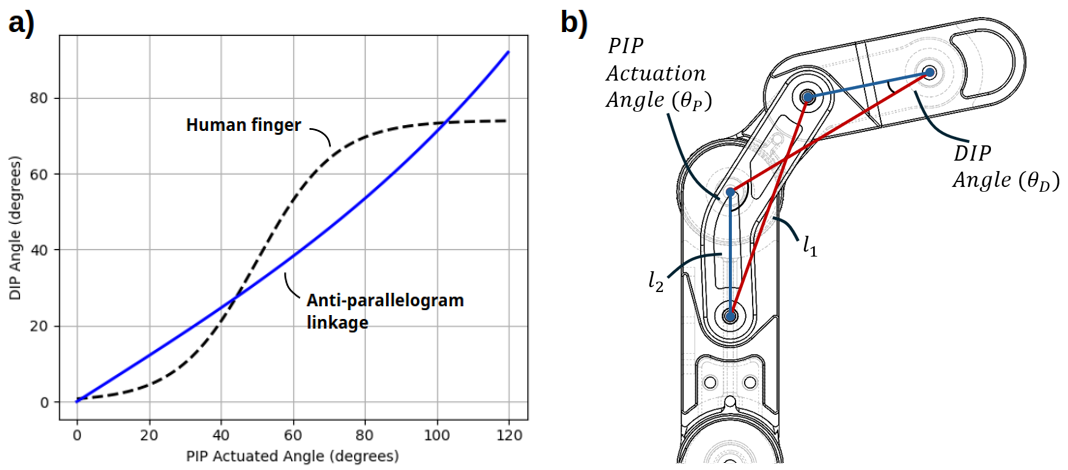}
    \caption{Configuration of PIP and DIP joints. a) Comparison of DvPT between the anti-parallelogram linkage and an analytical fit of the human finger \cite{van2015analytical}. b) Anti-parallelogram linkage on the BiDexHand’s PIP and DIP joints.} 
    \label{fig:DvPT_angle}
\end{figure}

The linkages can be completely 3D-printed and assembled easily. By tuning the lengths of the crossed and non-adjacent links, the linkage can produce a trajectory that approximates the DvPT of the human finger, as shown in Figure~\ref{fig:DvPT_angle}.

Given the anti-parallelogram linkage configuration, we can find the passive DIP angle, $\theta_D$, from the following equation:

\begin{equation}
    \theta_{D} = 2 \arctan\left(\frac{l_1 - l_2}{l_1 + l_2} \cot\left(\frac{\theta_{P}}{2}\right)\right)
    \label{eq:1}
\end{equation}

where $l_1$ and $l_2$ are the lengths of the crossed side and non-adjacent side of the anti-parallelogram, and $\theta_P$ is the actuated PIP angle.

\subsection{Thumb Design}

The robotic thumb, like its finger counterparts, consists of four functionally decoupled joints arranged in a serial configuration: MCP flexion/extension and abduction/adduction, followed by carpometacarpal (CMC) abduction/adduction and flexion/extension, listed from distal to proximal.

While the human thumb’s interphalangeal (IP) joint is not mechanically coupled to the MCP joint, their motions exhibit neuromuscular coupling through motor synergies \cite{santello1998postural}. For instance, during tasks such as pinching or power grasping, both joints tend to flex in coordination. To capture this functional behavior and reduce actuator count, the robotic thumb incorporates an anti-parallelogram linkage similar to that of the fingers to passively couple IP and MCP flexion. Due to spatial constraints within the base of the thumb, the CMC flexion/extension joint was implemented using a compact parallelogram four-bar linkage.

\subsection{Cable-Driven Actuation}

Like human fingers, which are controlled by flexor and extensor muscles through tendons, BiDexHand utilizes a cable-driven mechanism. Servo motors are mounted in the proximal servo sleeve region, effectively relocating joint inertia away from the fingers and toward the base of the robot. Each active joint is driven by a dedicated servo via an N-configuration tendon routing scheme \cite{tendon_types}, in which one actuator controls a single degree of freedom. Endless tendon loops are used to create an antagonistic pull-pull configuration.

\begin{figure}[b]
    \centering
    \includegraphics[width=0.45\textwidth]{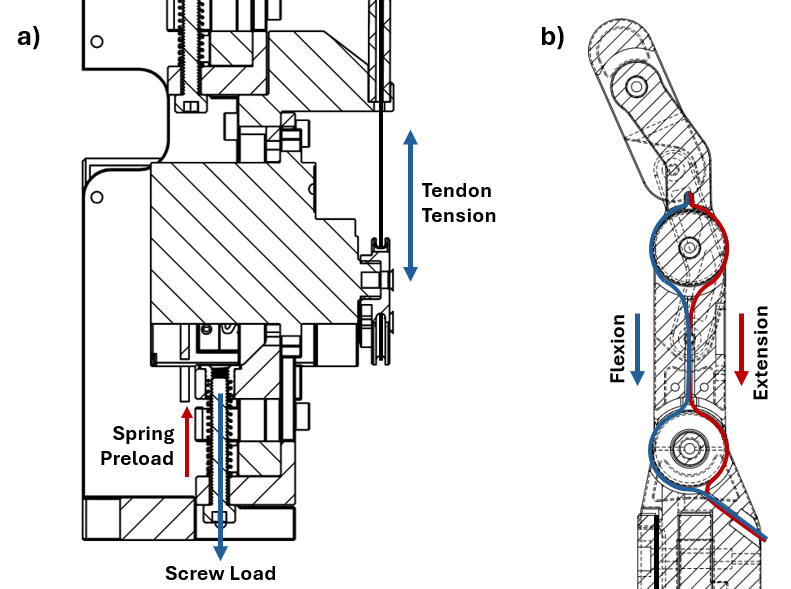}
    \caption{Overview of the cable-driven mechanism on BiDexHand. a) Tensioning mechanism in the servo sleeve region. b) Cable routing configuration for the robot finger’s PIP joint.}
    \label{fig:routing}
\end{figure}

The N-configuration routing offers several advantages while introducing some challenges. It requires the fewest actuators among common schemes, reducing system weight, cost, and mechanical complexity. With each DoF controlled independently, it minimizes the number of cables and pulleys, simplifying integration and dynamic modeling. However, since each tendon spans a fixed length from actuator to joint, proper pre-tensioning is essential to prevent slack and hysteresis, which can degrade control accuracy. To address this, a spring-loaded tensioning mechanism is integrated into the servo sleeve, allowing for continuous post-routing adjustment and ensuring reliable joint performance, as shown in Figure~\ref{fig:routing}.

PTFE tubes guide the tendons’ paths, offering enough flexibility to follow joint actuation. They also provide some preload to the tendons; although this increases friction, it ensures that the tendons maintain a good amount of tension. This design minimizes tendon routing inside the linkages and greatly reduces the mechanical complexity of the robotic hand.

\begin{figure*}[!b]
    \centering
    \includegraphics[width=0.95\textwidth]{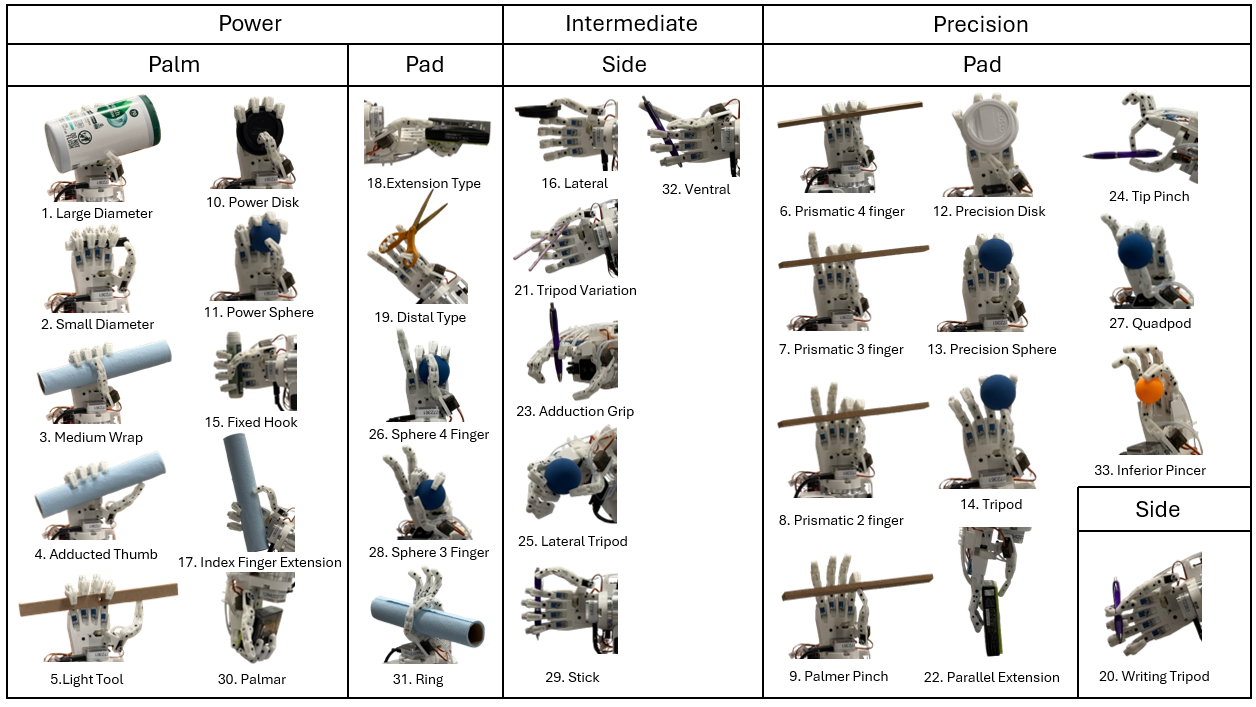}
    \caption{Assessing the GRASP taxonomy on the BiDexHand. The robotic hand achieves 33 out of 33 grasp poses.}
    \label{fig:grasp_tax}
\end{figure*}

\subsection{Control Architecture}
Position control was implemented on the robotic hand, with the control software developed based on the ROS2 framework. Joint angle commands are transmitted via serial communication to an ESP32 microcontroller onboard, which drives the servo motors in the tendon-drive mechanism. To accommodate different operational requirements, three primary control modes were developed: joint-level control, task-based control, and motion shadowing. In the motion shadowing mode, hand landmarks are detected through vision-based inputs, either from an RGB camera or a VR headset. Hand joint angles are calculated from the landmarks, then mapped onto servo joint angles. Although the current implementation relies on open-loop position control, development of a closed-loop solution is underway.

\section{Dexterity Evaluation}

Evaluating the performance of a dexterous robotic hand is essential for validating its suitability in real-world manipulation tasks. Key metrics such as range of motion, functionality, coordination, and object interaction help determine whether the hand meets its intended design objectives. While various methodologies have been proposed in the literature to quantify robotic dexterity \cite{wang2017thumb}\cite{zhou202050}, this work focuses on two representative evaluation strategies that together provide a broad assessment of the BiDexHand’s capabilities:

\begin{itemize}
    \item \textbf{Grasp Taxonomy:} Based on the 33-class taxonomy by Feix et al. \cite{grasp_tax}, this evaluation tests BiDexHand’s ability to replicate a wide range of human grasps, assessing its adaptability across various object shapes, sizes, and materials.

    \item \textbf{Thumb Dexterity:} Using the Kapandji score \cite{kapandji1986clinical}, thumb opposability is evaluated by measuring its ability to reach key positions on the opposing fingers. This metric reflects the effectiveness of the thumb design for fine manipulation.
\end{itemize}

Together, these evaluation methods provide a comprehensive view of the BiDexHand’s dexterous performance. The following sections detail the experimental procedures and present results that highlight both the capabilities and limitations of the current design.

\subsection{Grasp Taxonomy}

The BiDexHand’s multi-DoF coordination and range of motion were evaluated using the GRASP Taxonomy benchmark introduced by Feix et al. \cite{grasp_tax}. The hand was tasked with performing all 33 defined grasp types while securely holding appropriate objects to meet the criteria for each pose. Figure~\ref{fig:grasp_tax} shows the grasp types performed during the evaluation.

As the result of the evaluation, the BiDexHand successfully completed all 33 grasps and achieved 9 out of 11 Kapandji Score, demonstrating a high degree of dexterity and functional versatility across a wide range of object geometries and manipulation contexts.

\subsection{Thumb Dexterity}

\begin{figure}[h]
    \centering
    \includegraphics[width=0.475\textwidth]{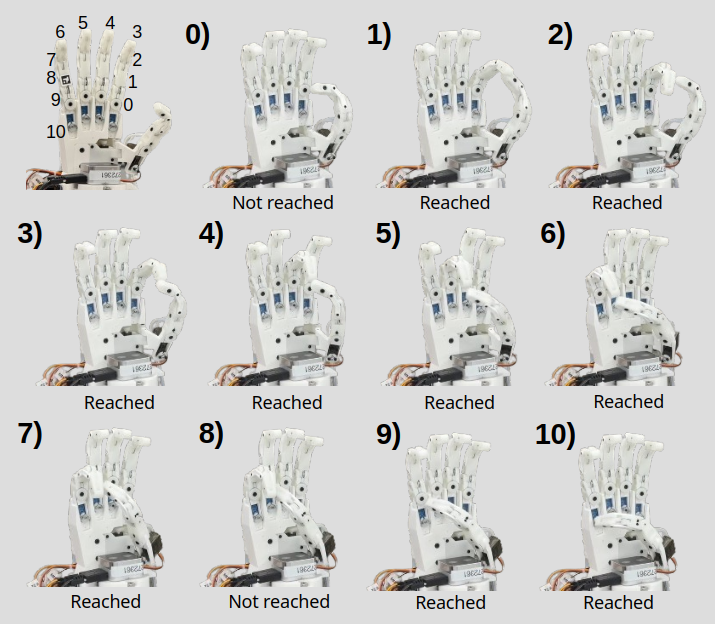}
    \caption{Assessing Kapandji score on the BiDexHand. The hand achieves 9 out of 11 poses. Poses 0 and 8 were not reached due to the CMC joint’s range of motion.}
    \label{fig:hand_kapandji}
\end{figure}

To assess thumb opposability and range of motion, the Kapandji test \cite{kapandji1986clinical} was conducted on the BiDexHand. The robotic hand was commanded to specific joint positions to achieve the poses described in the evaluation metric, creating point contact between the linkages. As a result, the BiDexHand was able to reach 9 out of 11 reference positions, indicating effective thumb articulation and opposition for most in-hand tasks. Figure~\ref{fig:hand_kapandji} illustrates the thumb positions evaluated in the Kapandji test.

The two unreachable Kapandji poses (\#0 and \#8) were limited by the thumb’s carpometacarpal (CMC) design, which uses two decoupled serial joints rather than a single composite joint. This constraint reduced the thumb’s ability to fully reach the extremes of opposition range, with \#0 close to the root of the index finger and \#8 near the PIP joint of the little finger.

\section{Force Evaluation}
\begin{figure}[t]
    \centering
    \includegraphics[width=0.475\textwidth]{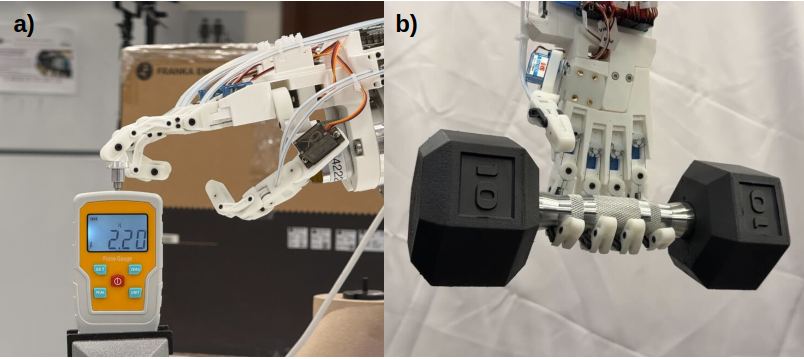}
    \caption{Assessing force output of the BiDexHand. a) The robotic finger exerts a fingertip load of up to 2.2~N. b) The robotic hand lifts a 10~lb weight in a hooking gesture.}
    \label{fig:hand_force}
\end{figure}

In addition to evaluating dexterity, the force output capabilities of the BiDexHand were assessed. Fingertip force was measured using a calibrated sensor across five independent trials per finger, resulting in an average peak output of 2.14~N. The hand’s lifting capability was similarly evaluated by demonstrating a successful grasp and lift of a 10~lb (approximately 4.54~kg) dumbbell. Although the system was not specifically designed for high-force applications, these results highlight the mechanical robustness of the tendon-driven design and its potential for lifting heavier objects under whole-hand grasps. Figure~\ref{fig:hand_force} illustrates the robotic hand configurations used during the force tests.

\section{Open Source}
One of the key objectives of this work is to foster collaboration and accelerate advancements in dexterous manipulation. To that end, the full design files, control software, and documentation for the proposed biomimetic hand are made freely available in an open-source repository \cite{GitHubRepo}. By providing accessible hardware specifications and modular code, this platform enables researchers, students, and industry professionals to customize, extend, and build upon the system. Such collective innovation aims to drive forward the development of robust, low-cost dexterous hands for both academic and industrial applications.

\section{Conclusion and Future Work}

This work presents BiDexHand, a cable-driven, open-source biomimetic robotic hand that demonstrates high dexterity through a novel mechanical design. Evaluation using the GRASP Taxonomy and Kapandji score validates the system’s versatility: BiDexHand successfully performed all 33 standard grasp types and achieved 9 out of 11 reference positions in the thumb dexterity test. Additionally, the robotic hand is capable of delivering a fingertip force of 2.14~N and lifting a 10~lb weight. These results underscore the effectiveness of the hand’s tendon-driven design and joint architecture in enabling complex manipulation tasks.

Building on this strong foundation, future work will focus on extending the current open-loop controller to a closed-loop system using vision-based feedback, further optimizing tendon routing to reduce backlash and improve precision, and exploring learning-based strategies to enhance grasp planning and adaptation in unstructured environments.

\bibliographystyle{IEEEtran}
\bibliography{References}  

\begin{thebibliography}{10}
\providecommand{\url}[1]{#1}
\csname url@samestyle\endcsname
\providecommand{\newblock}{\relax}
\providecommand{\bibinfo}[2]{#2}
\providecommand{\BIBentrySTDinterwordspacing}{\spaceskip=0pt\relax}
\providecommand{\BIBentryALTinterwordstretchfactor}{4}
\providecommand{\BIBentryALTinterwordspacing}{\spaceskip=\fontdimen2\font plus
\BIBentryALTinterwordstretchfactor\fontdimen3\font minus \fontdimen4\font\relax}
\providecommand{\BIBforeignlanguage}[2]{{%
\expandafter\ifx\csname l@#1\endcsname\relax
\typeout{** WARNING: IEEEtran.bst: No hyphenation pattern has been}%
\typeout{** loaded for the language `#1'. Using the pattern for}%
\typeout{** the default language instead.}%
\else
\language=\csname l@#1\endcsname
\fi
#2}}
\providecommand{\BIBdecl}{\relax}
\BIBdecl

\bibitem{bridgwater2012robonaut}
L.~B. Bridgwater, C.~Ihrke, M.~A. Diftler, M.~E. Abdallah, N.~A. Radford, J.~Rogers, S.~Yayathi, R.~S. Askew, and D.~M. Linn, ``The robonaut 2 hand-designed to do work with tools,'' in \emph{2012 IEEE International Conference on Robotics and Automation}.\hskip 1em plus 0.5em minus 0.4em\relax IEEE, 2012, pp. 3425--3430.

\bibitem{kim2021ilda}
U.~Kim, D.~Jung, H.~Jeong, J.~Park, H.-M. Jung, J.~Cheong, H.~R. Choi, H.~Do, and C.~Park, ``Integrated linkage-driven dexterous anthropomorphic robotic hand,'' \emph{Nature communications}, vol.~12, no.~1, p. 7177, 2021.

\bibitem{shadowhand_patent}
A.~De~La Rosa~Tames, G.~R.~L. Walker, J.~B. Goldsmith, J.~H. Elias, M.~P. Godden, and R.~M. Greenhill, ``Robotic hand,'' Patent or Publication Details, Feb. 2014, current Assignee: Shadow Robot Co Ltd, Filed June 25, 2010, issued February 25, 2014. Available at: \url{https://patents.google.com/patent/US8660695B2/en}.

\bibitem{shadowrobot_hand_cost}
\BIBentryALTinterwordspacing
{Shadow Robot}, ``How much does a robot hand cost?'' n.d., accessed: 7 April 2025. [Online]. Available: \url{https://www.shadowrobot.com/blog/how-much-does-a-robot-hand-cost/}
\BIBentrySTDinterwordspacing

\bibitem{hand_dof}
J.~Lee and T.~Kunii, ``Model-based analysis of hand posture,'' \emph{IEEE Computer Graphics and Applications}, vol.~15, no.~5, pp. 77--86, 1995.

\bibitem{LEIJNSE20102381PIPKINEMATIC}
\BIBentryALTinterwordspacing
J.~Leijnse, P.~Quesada, and C.~Spoor, ``Kinematic evaluation of the finger’s interphalangeal joints coupling mechanism—variability, flexion–extension differences, triggers, locking swanneck deformities, anthropometric correlations,'' \emph{Journal of Biomechanics}, vol.~43, no.~12, pp. 2381--2393, 2010. [Online]. Available: \url{https://www.sciencedirect.com/science/article/pii/S0021929010002332}
\BIBentrySTDinterwordspacing

\bibitem{van2015analytical}
K.~J. Van-Zwieten, K.~P. Schmidt, G.~J. Bex, P.~L. Lippens, and W.~Duyvendak, ``An analytical expression for the dip--pip flexion interdependence in human fingers,'' \emph{Acta of Bioengineering and Biomechanics}, vol.~17, no.~1, pp. 129--135, 2015.

\bibitem{santello1998postural}
M.~Santello, M.~Flanders, and J.~F. Soechting, ``Postural hand synergies for tool use,'' \emph{Journal of neuroscience}, vol.~18, no.~23, pp. 10\,105--10\,115, 1998.

\bibitem{tendon_types}
H.~Sang, C.~Yang, F.~Liu, J.~Yun, and F.~Chen, ``Kinematics rapid modeling method of tendon-driven robotic mechanisms using the tendon-routing matrix and equivalent radius matrix,'' in \emph{2016 International Conference on Electrical, Mechanical and Industrial Engineering}.\hskip 1em plus 0.5em minus 0.4em\relax Atlantis Press, 2016, pp. 300--303.

\bibitem{wang2017thumb}
H.~Wang, S.~Fan, and H.~Liu, ``Thumb configuration and performance evaluation for dexterous robotic hand design,'' \emph{Journal of Mechanical Design}, vol. 139, no.~1, p. 012304, 2017.

\bibitem{zhou202050}
J.~Zhou, Y.~Chen, D.~C.~F. Li, Y.~Gao, Y.~Li, S.~S. Cheng, F.~Chen, and Y.~Liu, ``50 benchmarks for anthropomorphic hand function-based dexterity classification and kinematics-based hand design,'' in \emph{2020 IEEE/RSJ International Conference on Intelligent Robots and Systems (IROS)}.\hskip 1em plus 0.5em minus 0.4em\relax IEEE, 2020, pp. 9159--9165.

\bibitem{grasp_tax}
T.~Feix, J.~Romero, H.-B. Schmiedmayer, A.~M. Dollar, and D.~Kragic, ``The grasp taxonomy of human grasp types,'' \emph{IEEE Transactions on Human-Machine Systems}, vol.~46, no.~1, pp. 66--77, 2016.

\bibitem{kapandji1986clinical}
A.~Kapandji, ``Clinical test of apposition and counter-apposition of the thumb,'' \emph{Annales de chirurgie de la main: organe officiel des societes de chirurgie de la main}, vol.~5, no.~1, pp. 67--73, 1986.

\bibitem{GitHubRepo}
Z.~K. Weng, ``{BiDexHand: An Open-Source Biomimetic Dexterous Hand},'' \url{https://github.com/wengmister/BiDexHand}, 2025.

\end{thebibliography}

\end{document}